 \documentclass[final,5p,times,twocolumn]{elsarticle}

\usepackage[T1]{fontenc}
\usepackage[utf8]{inputenc}
\usepackage{graphicx}
\usepackage{xcolor}

\usepackage{bm}
\usepackage{amsmath}
\usepackage{amssymb}
\usepackage{amsfonts}
\usepackage{thmtools}		
\usepackage{mleftright}
\usepackage{stmaryrd}
\usepackage{nicefrac}
\usepackage{algorithm}
\usepackage{algorithmicx}
\usepackage{subfigure}
\usepackage{hyperref}
\usepackage{svg}

\newcommand{\xt}{\mathbf{x}^{(t)}}
\newcommand{\yt}{\mathbf{y}^{(t)}}
\newcommand{\xtt}{\mathbf{x}^{(t+1)}}
\newcommand{\ytt}{\mathbf{y}^{(t+1)}}
\newcommand{\ioft}{\mathbf{i}^{(t)}}
\newcommand{\jt}{\mathbf{j}^{(t)}}

\newcommand{\W}{\mathbf{W}}
\newcommand{\Wrr}{\mathbf{W}_r}
\newcommand{\Wi}{\mathbf{W}_i}
\newcommand{\Wout}{\mathbf{W}_{\mathrm{out}}}

\newcommand{\w}{\mathbf{w}}
\newcommand{\x}{\mathbf{x}}
\newcommand{\y}{\mathbf{y}}
\newcommand{\be}{\mathbf{e}}
\newcommand{\bi}{\mathbf{i}}
\newcommand{\bj}{\mathbf{j}}
\newcommand{\bo}{\mathbf{o}}
\newcommand{\bu}{\mathbf{u}}
\newcommand{\but}{\mathbf{u}^{(t)}}
\newcommand{\bvt}{\mathbf{v}^{(t)}}
\newcommand{\bv}{\mathbf{v}}
\newcommand{\bG}{\mathbf{G}}

\begin{document}

\begin{frontmatter}

\title{Comparison of Reservoir Computing topologies using the Recurrent Kernel approach}

\author[first]{Giuseppe Alessio D'Inverno}
\affiliation[first]{organization={Department of Information Engineering and Mathematics, University of Siena},
            city={Siena},
            postcode={53100}, 
            country={Italy}}
\author[second]{Jonathan Dong}
\affiliation[second]{organization={Biomedical Imaging Group, École Polytechnique Fédérale de Lausanne},
            addressline={Station 17}, 
            city={Lausanne},
            postcode={1015}, 
            country={Switzerland}}

\begin{abstract}

Reservoir Computing (RC) has become popular in recent years thanks to its fast and efficient computational capabilities. Standard RC has been shown to be equivalent in the asymptotic limit to Recurrent Kernels, which helps in analyzing its expressive power. However, many well-established RC paradigms, such as Leaky RC, Sparse RC, and Deep RC, are yet to be systematically analyzed in such a way. We define the Recurrent Kernel limit of all these RC topologies and conduct a convergence study for a wide range of activation functions and hyperparameters. Our findings provide new insights into various aspects of Reservoir Computing. First, we demonstrate that there is an optimal sparsity level which grows with the reservoir size. Furthermore, our analysis suggests that Deep RC should use reservoir layers of decreasing sizes. Finally, we perform a benchmark demonstrating the efficiency of Structured Reservoir Computing compared to vanilla and Sparse Reservoir Computing.

\end{abstract}


\end{frontmatter}



\section{Introduction and related work}\label{sec:intro}

Reservoir Computing (RC) is a machine learning technique used for training Recurrent Neural Networks, which fixes the internal weights of the network and only trains a linear layer, resulting in faster training times~\cite{jaeger2001echo}. Its simplicity and effectiveness have made it a popular choice for various tasks such as chaotic time series prediction \cite{jaeger2001echo}, robot motor control or financial forecasting~\cite{lukovsevivcius2009reservoir}. Additionally, the random connections within Reservoir Computing networks make them a useful framework for comparison with biological neural networks \cite{damicelli2022brain}. 

Over time, researchers have proposed several methods to optimize and enhance Reservoir Computing's performance and efficiency.  
One such method is Leaky Reservoir Computing, which stabilizes the dynamics of the reservoir and enables tuning of its memory by adjusting the leak rate \cite{jaeger2007optimization}. There is also Sparse Reservoir Computing, which consists in a sparse initialisation of weight connections proposed since the original formulation of Echo State Networks \cite{jaeger2001echo}). 
Structured Reservoir Computing \cite{dong2020reservoir} is another acceleration strategy which replaces the internal weights by a structured transform instead. 
Finally, Deep Reservoir Computing allows for the use of reservoirs with different time dynamics \cite{gallicchio2017deep}. 
All these variants show the flexibility of the Reservoir Computing framework, made of a fixed reservoir encoding a time-dependent input combined with a linear readout, which can be extended further to physical implementations \cite{tanaka2019recent, dong2019optical, rafayelyan2020large} and next-generation reservoir computing \cite{gauthier2021next}.

Increasing the number of neurons in a Reservoir Computing network leads to the convergence of its behavior to a recurrent kernel, as discussed in \cite{dong2020reservoir}. Kernel methods are a class of algorithms in machine learning that use kernel functions to implicitly map input data into high-dimensional feature spaces, calculating scalar products between input points in a dual space, enabling linear models to solve non-linear problems.
Recurrent Kernels are a variant in which these scalar products are dynamically updated over time based on changes in the input data. They can be used as an alternative to large-scale Reservoir Computing and have demonstrated state-of-the-art performance on chaotic time series prediction \cite{dong2020reservoir}.
There are two ways Recurrent Kernels can be useful. 
They offer an interesting alternative to RC when the number of data points is limited, as kernel methods require the calculation of scalar products between all pairs of input points.  
Additionally, recurrent kernels have been useful for theoretical studies, such as stability analysis in Reservoir Computing \cite{dong2022asymptotic}, as they provide a deterministic limit with analytical expressions.

Prior studies on Recurrent Kernels has been mainly limited to vanilla Reservoir Computing and structured transforms. In this work, we extend the application of Recurrent Kernels to other Reservoir Computing topologies, such as Leaky, Sparse, and Deep Reservoir Computing. Specifically, we define the appropriate Recurrent Kernels for each topology, investigate similarities in their corresponding limits, and evaluate their convergence numerically. By broadening the scope of Recurrent Kernels, we aim to demonstrate their versatility and effectiveness in a range of Reservoir Computing configurations. 

Our main contributions are listed as follows: 
\begin{itemize}
    \item We define the Recurrent Kernel limit for Leaky RC, Sparse RC, and Deep RC, showing that Sparse RC converges to the same limit as vanilla RC and Structured RC
    \item We conduct a thorough numerical study on the convergence of these RC paradigms to their Recurrent Kernel counterparts, for different activation functions
    \item Our results show that sparse RC is equivalent to the non-sparse case, as long as the sparsity rate is above a certain threshold. This suggests that sparse RC does not have increased or decreased expressivity compared to vanilla RC
    \item We show that, in Deep Reservoir Computing, first reservoirs should be larger than subsequent ones, in order to decrease the amount of noise transmitted in the subsequent layers. However, this effect is quite small for large reservoirs and reservoirs with equal sizes should perform similarly in practice.
    \item We perform a benchmark of Reservoir Computing, Sparse Reservoir Computing, and Structured Reservoir Computing, as both strategies have been introduced for computational efficiency, and demonstrate that Structured RC is generally the most efficient for large reservoir sizes.
\end{itemize}

\section{Background}\label{sec:background}

\subsection{Reservoir Computing}

Reservoir Computing, like all recurrent neural network architectures, receives sequential input $\ioft \in \mathbb{R}^d$ for $t \in \mathbb{N}$. The simplest model for Reservoir Computing, commonly called the Echo-State Network (ESN) \cite{jaeger2004harnessing}, comprises a set of neurons $\xt \in \mathbb{R}^N$ with fixed random weights, with $N$ the number of neurons in the reservoir. The initial state of the network $\x^{(0)}$ is randomly initialized, typically from a random gaussian distribution. The network is then updated according to the following equation: 
\begin{equation}
    \label{eq: RC update}
    \xtt = \frac{1}{\sqrt{N}} f\left(  \sigma_r \Wrr \xt + \sigma_i \Wi \ioft \right).
\end{equation}
Here, $\Wrr \in \mathbb{R}^{N \times N}$ and $\Wi \in \mathbb{R}^{N \times d}$ are the reservoir and input weight matrices, $\sigma_r$ and $\sigma_i$ are reservoir and input scaling factors, and $f$ is an element-wise nonlinearity, often a sigmoid---which is well approximated by the (Gauss) error function. The factor $1/\sqrt{N}$ ensures proper normalization of the L2-norm of $\x$ when $N$ goes to infinity. Each weight of $\Wrr$ and $\Wi$ is drawn from a  normal Gaussian distribution with unit variance:
\begin{equation}
    \label{eq: initial gaussian weights}
    p(w) = \frac{1}{\sqrt{2\pi}} e^{-w^2/2}
\end{equation}

An essential hyperparameter to tune is the scaling factor $\sigma_r$. It significantly impacts the dynamics and stability of the reservoir:
when $\sigma_r$ is small, the updates in Eq. \eqref{eq: RC update} are contractant, while the reservoir becomes a chaotic nonlinear system for large $\sigma_r$. Therefore, this hyperparameter is often optimized to maximize performance for a given task. 

The output of Reservoir Computing $\bo^{(t)} \in \mathbb{R}^n$ is computed using a linear model applied to the state of the reservoir, as given by:
\begin{equation}\label{eq: linear output}
    \bo^{(t)} = \Wout \xt.
\end{equation}
The training step for this model involves a linear regression, which is a stark contrast to the non-linear optimization typically employed when training neural networks. Reservoir Computing's approach is based on the idea that the current state of the reservoir, $\xt$, non-linearly encodes the past values of the input time series, $\bi^{(t-1)}, \bi^{(t-2)},$ etc. 

\subsection{Different variants of Reservoir Computing}

\begin{figure*}
    \centering    
    \includegraphics[width=0.9\linewidth]{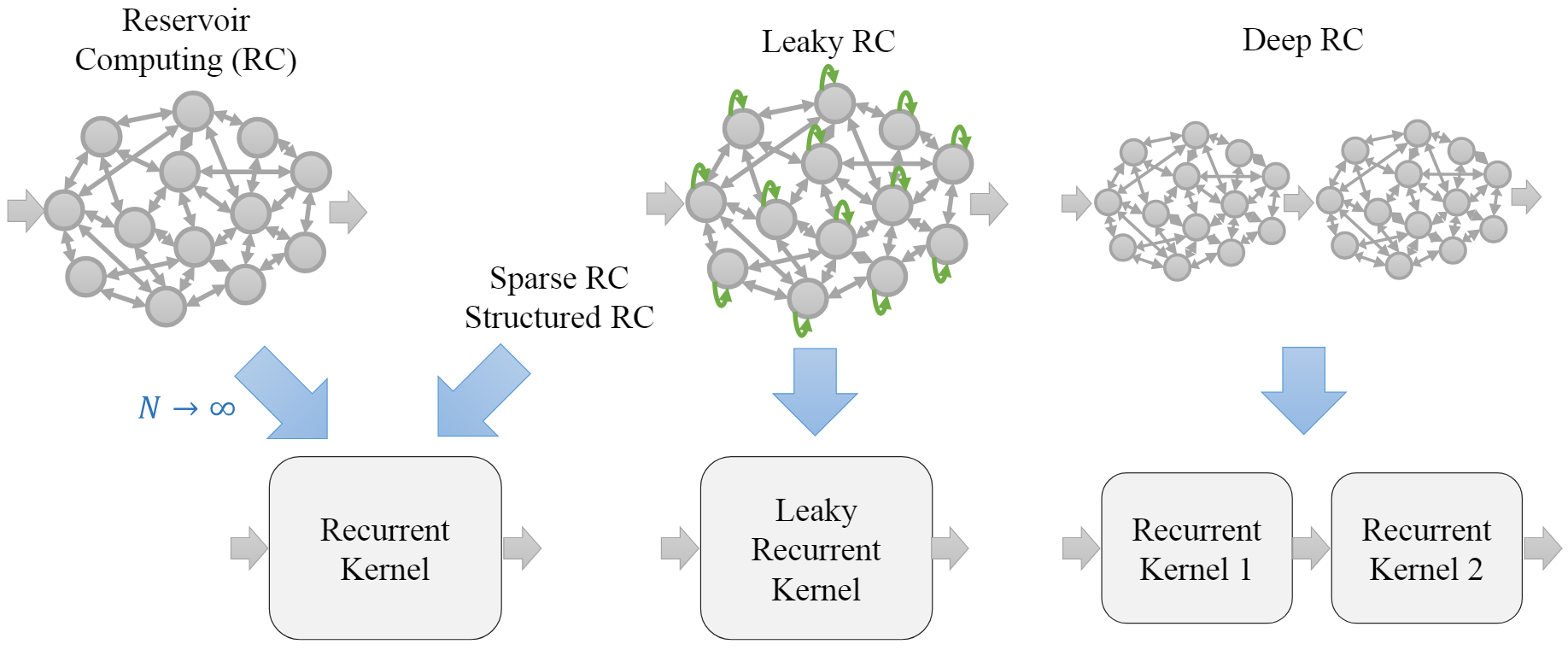}
    \caption{Recurrent Kernels associated with various Reservoir Computing topologies. RC, sparse RC, and structured RC converge to the same RK limit when the reservoir size $N \rightarrow \infty$. Leaky RC and Deep RC converge to their corresponding limits.}
    \label{fig:fig1}
\end{figure*}

Several Reservoir Computing variants have been proposed, which modify the update equations and alter the reservoir dynamics. These variants include adjustments to the updates to tune the reservoir relaxation time, speeding up computation, or introducing a hierarchical structure to enrich the dynamics. The flexibility of Reservoir Computing makes it possible to fine-tune the dynamics precisely for a particular task using these variants.

\textbf{Sparse Reservoir Computing} aims to increase computational efficiency by using sparse internal weight matrices. The computational complexity in Reservoir Computing is mainly determined by the matrix multiplication involving the $N \times N$ internal weight matrix. In Sparse Reservoir Computing, this matrix is made sparse by drawing the weights from a sparse i.i.d. distribution. Specifically, the distribution is given by:
\begin{equation}
    p(w_r) = (1-s) \delta(w_r) + s\sqrt{\frac{s}{2\pi}} e^{-\frac{s w_r^2}{2}},
\end{equation}
where $\delta$ denotes the Dirac delta function, and $s$ takes values between 0 and 1, controlling the proportion of non-zero weights. $s=1$ corresponds to the original non-sparse case, and it is typically set at 0.05 \cite{xue2007decoupled, gallicchio2011architectural} which means that 5\% of the weights are non-zero. The variance of the gaussian term is fixed at $1/s$ to ensure that the spectral radius of the matrix stays similar to the non-sparse case. We focus here on a sparsity model in which a fraction of the weights are non-zero. Other works define sparsity with a fixed number of connections per neurons \cite{gallicchio2020sparsity}, the two approaches being equivalent at fixed reservoir sizes. 

The sparsity in the weight matrix reduces the computational complexity of the update equation, enabling faster computation without sacrificing performance. The computational and memory complexities are $\mathcal{O}(s N^2)$. This approach is especially useful for large-scale Reservoir Computing systems where the computational cost is significant. Research has shown that the use of sparse matrix multiplication can increase computational speed, while maintaining accuracy. Thanks to their simplicity, they are often used in existing Reservoir Computing works. 

\textbf{Structured RC} speeds up computations by replacing the dense weight matrices by a product of fixed structured and random diagonal matrices \cite{dong2020reservoir}, inspired by Orthogonal Random Features \cite{yu2016orthogonal} which are their non-recurrent counterparts. The weight matrix $W$ is replaced by:
\begin{equation*}
    W = H D_1 H D_2 H D_3,
\end{equation*}
where $H$ denotes a fixed structured transform and $D_i$ are diagonal matrices with i.i.d. Rademacher random variables ($\pm 1$ with probability 0.5) on the diagonals. The computational complexity is $\mathcal{O}(n \log n)$, determined by the structured transforms, while the memory complexity is $\mathcal{O}(n)$. Deviating from \cite{dong2020reservoir}, we will consider Hartley matrices instead of Hadamard matrices. The Hartley transform is the real-valued version of the Fourier transform, defined as $H(x) = \operatorname{Re}(F(x*(1+j))$. It maps real-valued vectors to real-valued outputs and can be computed efficiently on CPU and GPU using the Fast Fourier Transform which is present in all generic numerical libraries in Python. By contrast, the Hadamard transform is a simple structured transform but requires dedicated Python libraries.

\textbf{Leaky-Reservoir Computing} introduces a leak rate to control the typical time scale of changes in the reservoir. The update equation for Leaky-Reservoir Computing is given by:
\begin{equation}
\xtt = (1-a) \xt + a \frac{1}{\sqrt{N}} f \left( \sigma_r \Wrr \xt + \sigma_i \Wi \ioft\right),
\label{eq: leaky RC definition}
\end{equation}
where $a \in [0,1]$ is the leak rate. Setting $a=1$ corresponds to the non-leaky Reservoir Computing case described earlier. Decreasing $a$ slows down the speed of changes in the reservoir, thereby controlling the typical time scale of reservoir dynamics. This feature can be useful for tasks in which the input signal changes slowly over time, as it enables the reservoir to better capture the temporal dependencies in the input data.

\textbf{Deep Reservoir Computing} stacks multiple reservoir layers to form a deep architecture also called a Deep Echo State Network (deepESN) \cite{gallicchio2017deep}. The first layer operates like the reservoir in a shallow Reservoir Computing architecture and is fed by the external input, while each successive layer is fed by the output of the previous one. The reservoir layer of a deepESN can be expressed as:
\begin{equation}
    \x_{l}^{(t+1)} = \frac{1}{\sqrt{N_l}} f \left( \sigma_r \Wrr^{l} \x_{l}^{(t)} + \sigma_i \Wi^{l} \bu_{l}^{(t)} \right)
\end{equation}
where the index $l = 1, \ldots, L$ describes the layer with a reservoir of size $N_l$, $\bu_{l}^{(t)}$ is the input for the $l$-th layer: 
\begin{equation}
    \bu_l^{(t)} = 
    \begin{cases}
        \bi^{(t)} \text{ if } l=1 \\
        \x_{l-1}^{(t+1)} \text{ if } l>1.
    \end{cases}
\end{equation}

One of the main ideas is that each reservoir is encoding the recent past of its received input. Thus, the first layer has limited memory while the subsequent ones are able to extend this memory and build more complex representations of the input signal. 

The output of a deepESN at each time step $t$ can be computed by applying any linear model to the different reservoir states. A common choice is to define the linear model on the concatenation of all reservoir states; the output $\bo^{(t)}$ is given by:
\begin{equation}
\bo^{(t)} = \Wout 
    \left[
    \begin{matrix}
        \x_{1}^{(t)} \\
        \x_{2}^{(t)} \\
        \vdots \\
        \x_{L}^{(t)}
    \end{matrix} \right]
\end{equation}
where $\Wout \in \mathbb{R}^{n \times \sum_l N_l}$ is a weight matrix that maps the concatenated reservoir states to the output. The concatenation of the reservoir states from each layer allows for the capture of information across multiple time scales, enabling the deepESN to model more complex temporal patterns.

The different variants presented above are not exclusive. For example, it is common to introduce different leak rates and sparse internal weight matrices to each layer of a deepESN. 

\textbf{Other strategies} have also been proposed to alleviate the dense matrix multiplication by the reservoir weights. Next-generation Reservoir Computing \cite{gauthier2021next} replaces the non-linear recurrent reservoir by an explicit mapping with polynomial combination of past inputs. As such, it can be interpreted as a non-recurrent temporal kernel. The flexibility of Reservoir Computing is further exemplified by the many physical implementations of Reservoir Computing \cite{tanaka2019recent, dong2019optical,rafayelyan2020large,huang2022prospects}, showing that any non-linear dynamical system can be used as a reservoir.

\subsection{Recurrent Kernels}

In machine learning, kernels are functions that measure the similarity between pairs of data points, in a high-dimensional space to enable effective linear models. This mapping into the higher-dimensional feature space is often done implicitly by computing the scalar products between data points. This observation can be extended to Reservoir Computing leading to Recurrent Kernels. We consider two reservoirs $\x$ and $\y$ driven by the inputs $\bi$ and $\bj$ respectively, following the update equation \eqref{eq: RC update}. For conciseness, we assume $\sigma_r = \sigma_i = 1$; equations for different values of reservoir and input scales can be obtained by substituting $\xt$ by $\sigma_r \xt$ and $\ioft$ by $\sigma_i \ioft$. The scalar product between two reservoir states can be expressed as:
\begin{align}
    \label{eq: explicit scalar product}
    \left(\xtt\right)^\top \ytt &= \frac{1}{N} \sum_{l = 1}^N f \left(\w_{r,l}^\top \xt + \w_{i,l}^\top \ioft \right) \times \nonumber \\
    &\qquad\qquad f \left( \w_{r,l}^\top \yt + \w_{i,l}^\top \jt \right)
\end{align}
where $\w_{r,l}$ and $\w_{i,l}$ denote the $l$-th line of $\Wrr$ and $\Wi$ respectively. Thanks to the law of large numbers, this quantity converges when the reservoir size $N$ goes to infinity to a deterministic kernel function $k_0:\left(\mathbb{R}^{N+d}\right)^2 \rightarrow \mathbb{R}$ defined as:
\begin{align}
    \label{eq: first kernel limit}
    k_0\left( \but, \bvt
    \right) 
    = \int d\w p(\w)  f\left( \w^\top \but \right) f\left( \w^\top \bvt \right)
\end{align}
where we have introduced $\but = \left[\begin{matrix}
    \xt \\
    \ioft
\end{matrix}\right]$, $\bvt = \left[\begin{matrix}
    \yt \\
    \jt
\end{matrix}\right]$, and $\w = \left[\begin{matrix}
    \w_r \\
    \w_i
\end{matrix}\right]$ a random vector of dimension $N+d$ with i.i.d. normal entries.

To properly define the associated Recurrent Kernel, we need to remove the dependency on the previous reservoir states $\xt$ and $\yt$, as they themselves depend on the random weights $\Wrr$ and $\Wi$. This is possible if $k_0$ is an \textit{iterable kernel}, i.e. if there exists $k:\mathbb{R}^3\rightarrow\mathbb{R}$ such that for all $\bu, \bv \in \mathbb{R}^{N+d}$: 
\begin{align}
    \label{eq: RK previous scalar products}
    k_0(\bu, \bv) = k(\|\bu\|^2, \|\bv\|^2, \bu^\top \bv).
\end{align}
We show in Appendix that the kernel associated to Reservoir Computing is always iterable, as soon as the weights are sampled from a gaussian distribution (or any rotationnally-invariant distribution $p(\w)$). This is an extension of the statement in \cite{dong2020reservoir} that assumed translation or rotation-invariant kernels. 

We define the recurrent kernel by replacing $\left(\xtt\right)^\top \ytt$ in Eq. \eqref{eq: explicit scalar product} by $k^{(t+1)}\left(\ioft, \jt, \ldots\right)$. Similarly, we replace in Eq. \eqref{eq: RK previous scalar products} $\left(\but\right)^\top \bvt$
by $k^{(t)}(\bi^{(t-1)}, \bj^{(t-1)}, \ldots) + \left(\ioft\right)^\top \jt$, and perform the same operation for the norms (as norms are symmetric scalar products). This leads to the following definition of a Recurrent Kernel (RK) as a sequence of kernel functions $k^{(t)}:\left(\mathbb{R}^d\right)^{2t} \rightarrow \mathbb{R}$ for $t \in \mathbb{N}^*$:
\begin{equation}
    \begin{cases}
        k^{(1)}\left(\bi^{(0)}, \bj^{(0)}\right) &= 
        k(1+ \|\bi^{(0)}\|^2, 1 + \|\bj^{(0)}\|^2, \left(\ioft\right)^\top \jt)
        \\
        k^{(t+1)}\left(\ioft, \jt, \ldots\right) &= 
        k(k^{(t)}(\bi^{(t-1)}, \bi^{(t-1)}, \ldots) + \|\ioft\|^2, \\
        &\qquad k^{(t)}(\bj^{(t-1)}, \bj^{(t-1)}, \ldots) + \|\jt\|^2, \\
        &\qquad k^{(t)}(\bi^{(t-1)}, \bj^{(t-1)}, \ldots) + \left(\ioft\right)^\top \jt)
    \end{cases}
    \label{eq: RK definition}
\end{equation}
In the first line, we have initialized the RK by choosing $\|\x^{(0)}\|=\|\y^{(0)}\|=1$ and $\left(\x^{(0)}\right)^\top \y^{(0)} = 0$. 

\begin{figure*}
    \centering
    \includegraphics[width=\linewidth]{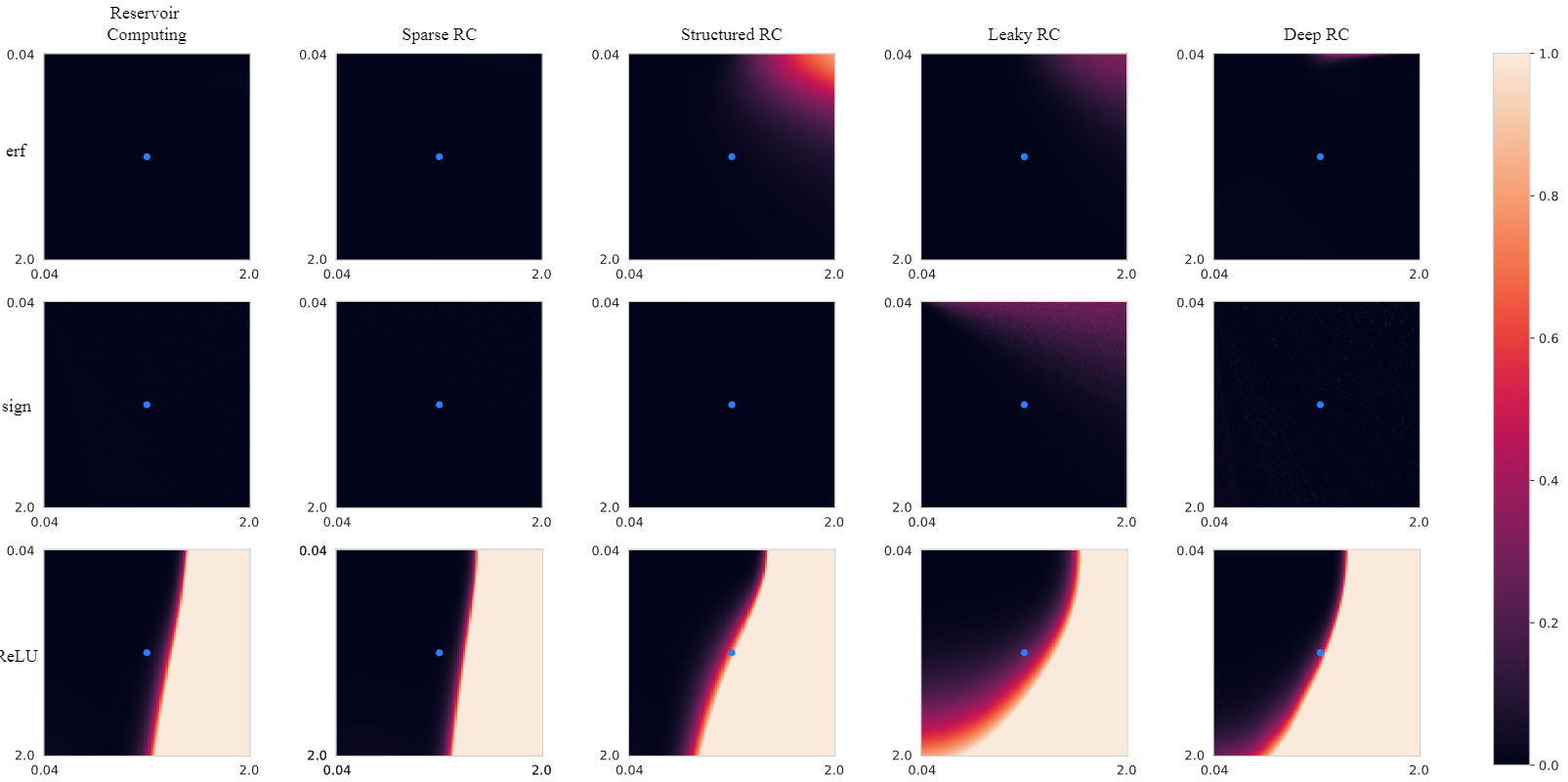}
    \caption{Convergence study of various Reservoir Computing topologies (columns) towards their corresponding Recurrent Kernel limits, for different activation functions $f$ (rows). For each case, the Frobenius norm between the RC and RK Gram matrices $L$ (Eq. \eqref{eq: metric convergence}, smaller is better) is displayed for weight scaling factors $\sigma_r$, $\sigma_i$ between 0.04 and 2. Blue dot in the first row: typical operating point $\sigma_r = \sigma_i = 1$.}
    \label{fig: fig2 convergence}
\end{figure*}

One can then replace large-scale Reservoir Computing by Recurrent Kernels. To accomplish this, one must compute the recurrent kernels \textit{for each pair of training inputs}, which are then placed into a matrix known as the Gram matrix. For instance, let us denote by $\bi_m$, $m=1,\ldots,M$, the different inputs. The Gram matrix $\bG^{(t)} \in \mathbb{R}^{M \times M}$ is defined for $t \in \mathbb{N}^*$ as:
\begin{equation}
    \label{eq: Gram matrix def}
    \bG^{(t)} = \left[ k^{(t)}(\bi_n^{(t-1)}, \bi_m^{(t-1)}, \ldots) \right]_{n,m}
\end{equation}
A linear model is trained using the Gram matrix and can be employed for making predictions. 

Recurrent Kernels hold promise as a substitute for large-scale Reservoir Computing, as they offer comparable performance as the Reservoir Computing limit with infinitely-many neurons. However, the drawback is the computational time needed for prediction, as scalar products must be computed with each training input for use in the linear model. Additionally, due to their analytic formulation, Recurrent Kernels are well-suited for theoretical studies on Reservoir Computing \cite{dong2022asymptotic}.

Rigorously proving the convergence of Reservoir Computing to the Recurrent Kernel has proven challenging. Three assumptions are typically required \cite{dong2020reservoir}: 
\begin{enumerate}
    \item Lipschitz-continuity: the activation function is $l$-Lipschitz;
    \item Contractivity: the scaling factor of the reservoir weights needs to satisfy $\sigma_r^2 \leq 1 / l$
    \item Time-independence: the weight matrix is resampled at each time step.
\end{enumerate}
These assumptions are very restrictive to prove convergence of RC towards their Recurrent Kernel limits in practice. We instead resort to numerical investigations of convergence, as presented in the next section.

\section{Recurrent Kernel Limits for Various RC Topologies}\label{sec:rec_kernels}

Here we define the Recurrent Kernel limits of the different Reservoir Computing topologies and discuss the assumptions for convergence to these Recurrent Kernels. More details are provided in the Appendix. 

The Recurrent Kernel for sparse Reservoir Computing corresponds to the same Recurrent Kernel as the non-sparse case. This implies that the asymptotic performance of a sparse reservoir is equivalent to the one of a nonsparse one. To obtain this result, the reservoir activations at each iteration needs not to be sparse, which is generally valid for Reservoir Computing. A more detailed study of the sparse case is provided in Appendix. This is similar to Structured Reservoir Computing: both strategies have the same RK limit and they have been introduce to decrease the cost of RC computations. As such, all three topologies (vanilla, sparse, and structured RC) are equivalent asymptotically and could be used interchangeably.

The Recurrent Kernel corresponding to Reservoir Computing with leak rate is defined by replacing the update equation of Eq. \eqref{eq: RK definition}
\begin{align}
    \label{eq: leaky RK definition}
    k^{(t+1)}(\ioft, \jt, \ldots) =& \ (1-a)^2 k^{(t)}\left(\bi^{(t-1)}, \bj^{(t-1)}, \ldots\right) \nonumber \\
    +&a^2
    k(k^{(t)}(\bi^{(t-1)}, \bi^{(t-1)}, \ldots) + \|\ioft\|^2, \nonumber \\
    &\qquad k^{(t)}(\bj^{(t-1)}, \bj^{(t-1)}, \ldots) + \|\jt\|^2, \nonumber \\
    &\qquad k^{(t)}(\bi^{(t-1)}, \bj^{(t-1)}, \ldots) + \left(\ioft\right)^\top \jt)
\end{align}
More details are provided in Appendix. As described for the vanilla case of Recurrent Kernels, this limit is valid beyond these assumptions and we will study it numerically.

For Deep Reservoir Computing, we can write the kernel limit for each layer. We start by the first layer:
\begin{align}
    \label{eq: deep RK 1st layer definition}
    k^{(t+1)}_{1}\left(\ioft, \jt, \ldots\right) &= k(k^{(t)}_1(\bi^{(t-1)}, \bi^{(t-1)}, \ldots) + \|\ioft\|^2, \nonumber \\
    &\qquad k^{(t)}_1(\bj^{(t-1)}, \bj^{(t-1)}, \ldots) + \|\jt\|^2, \nonumber \\
    &\qquad k^{(t)}_1(\bi^{(t-1)}, \bj^{(t-1)}, \ldots) + \left(\ioft\right)^\top \jt)
\end{align} 
For the subsequent layers $l>1$, we obtain the recursive formula:
\begin{align}
    k^{(t+1)}_{l}\left(\ioft, \jt, \ldots\right) &= k(k^{(t)}_l(\bi^{(t-1)}, \bi^{(t-1)}, \ldots) + k^{(t+1)}_{l-1}(\ioft, \ioft, \ldots), \nonumber \\
    &\qquad k^{(t)}_l(\bj^{(t-1)}, \bj^{(t-1)}, \ldots) + k^{(t+1)}_{l-1}(\jt, \jt, \ldots), \nonumber \\
    &\qquad k^{(t)}_l(\bi^{(t-1)}, \bj^{(t-1)}, \ldots) + k^{(t+1)}_{l-1}(\ioft, \jt, \ldots))
\end{align}
In practice, we can compute these Recurrent Gram matrices layer by layer.

The Recurrent Kernel prediction is performed by concatenating the reservoir states of the different layers and using a linear model. This corresponds to a sum of the different Gram matrices:
\begin{equation}
    k_\mathrm{tot}^{(t+1)}(\ioft, \jt, \ldots) = \sum_l k_l^{(t+1)} (\ioft, \jt, \ldots)
\end{equation}

\section{Results}\label{sec:results}
\subsection{Convergence study}

We show in Figure \ref{fig: fig2 convergence} a numerical convergence study of the various Reservoir Computing topologies to their respective Recurrent Kernel limits. Two random inputs of length $T=10$ and dimension $d=100$ are generated and fed to reservoirs of size $N=1000$. When applicable, the sparsity level is set at $s=0.5$ and the leak rate at $a=0.5$. For Deep Reservoir Computing, we use a sequence of two reservoirs of size $N_1 = N_2 = 1000$. 

We compute the final Gram matrix $G^\mathrm{RC}$ at time $T+1=11$. This Gram matrix is compared with the Gram matrix $G^\mathrm{RK}$ obtained using the associated Recurrent Kernel. The metric displayed is the Frobenius norm between the final Gram matrices:
\begin{equation}
    \label{eq: metric convergence}
    L = \left\|G_\mathrm{RC}^{(T+1)} - G_\mathrm{RK}^{(T+1)}\right\|^2_F.
\end{equation}

This metric is computed for different values of the reservoir weight standard deviations $\sigma_r$ and $\sigma_i$ which dictate the dynamics of the reservoir between 0 and 2. When $\sigma_r$ is small, dynamics are contractant, while they become chaotic for large values of $\sigma_r$. Theory only predicts convergence for the contractant case but it has also been observed for large $\sigma_r$. We typically choose $\sigma_r$ close to 1 to obtain non-chaotic but rich reservoir dynamics. We perform this study for three typical activation functions: erf (differentiable and bounded), sign (discontinuous and bounded), and ReLU (sub-differentiable and unbounded). Among these three, the error function would be the one generally used in practice.

We see that in the sparse case, the convergence is fundamentally similar to non-sparse Reservoir Computing. We observe convergence over the whole parameter range for bounded activation functions. For ReLU, convergence is achieved for values of $\sigma_r$ below a threshold which depends slightly on $\sigma_i$. This is due to errors accumulating with the ReLU activation function, for which convergence is more difficult to achieve compared to bounded activation functions. Convergence is also achieved over a wide range of parameters for Structured RC; the convergence region is slightly smaller for erf and ReLU, but most importantly Structured RC converges to the RK limit for the typical operating point in blue with the erf activation function.

For leaky Reservoir Computing and Deep Reservoir Computing, the convergence region is also smaller. For bounded activations, it typically does not converge for large $\sigma_r$ and small $\sigma_i$. This typically corresponds to an unstable case \cite{dong2022asymptotic} that is not used in practice. For ReLU, convergence is achieved for a range of parameters $(\sigma_r, \sigma_i)$ slightly smaller than the non-leaky case. When the reservoir or input weights are large, error accumulates and activations diverge. 

In general, convergence is achieved for a wide range of parameters with bounded activation functions. Caution is necessary only when $\sigma_r$ is large and $\sigma_i$ is small. For the ReLU case, convergence is more challenging, as activations may diverge.

\subsection{How to choose sparsity level}

\begin{figure}
    \includegraphics[width=\linewidth]{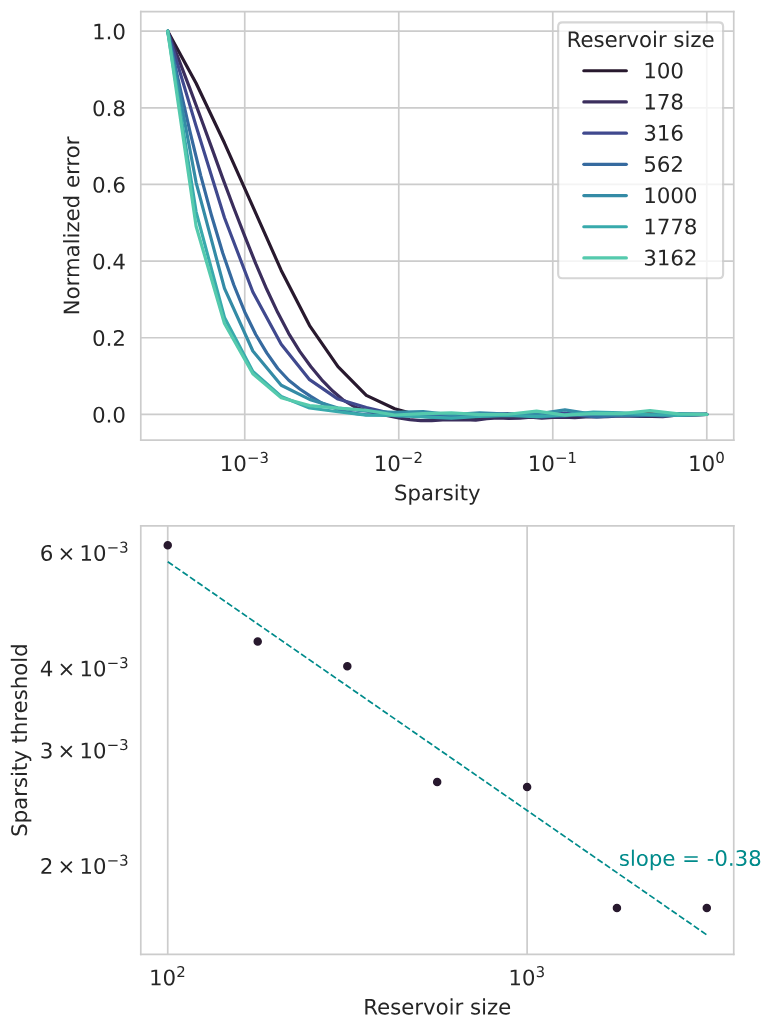}
    \caption{(Top) Error metric (Eq. \eqref{eq: metric convergence}) normalized between 0 and 1 as a function of sparsity for different reservoir sizes. (Bottom) Sparsity threshold above which the error metric is within $5\%$ of the non-sparse limit. This gives an admissible sparsity level which decreases with the reservoir size.}
    \label{fig: sparse}
\end{figure}

Our framework enables us to determine the optimal sparsity level to obtain the same convergence to the RK limit while decreasing the computational cost. We compute the normalized error metric as defined in Eq. \eqref{eq: metric convergence} for different reservoir sizes in Fig. \ref{fig: sparse} (top row). The error metric being dependent on the reservoir size, we normalize it between 0 and 1 to represent all curves on the same graph. The error metric being dependent on the stochastic realization of the weights, we perform a Monte-Carlo estimation with at least $10^4$ realizations to decrease the estimation variance. 

We observe that one can decrease the sparsity level until a threshold below which the approximation error increases. This threshold decreases with the reservoir size: larger reservoirs handle low sparsity levels better.

In the bottom row of Fig. \ref{fig: sparse}, we plot the threshold defined as $5\%$ of the non-sparse limit. We see that we can decrease the sparsity factor quite dramatically: for a reservoir size $N=1000$, we can decrease the sparsity rate down to $s=0.003$. This value is significantly below the typical sparsity level of 0.05 \cite{xue2007decoupled, gallicchio2011architectural}, which could lead to further computational and memory savings. 

This sparsity level corresponds to a mean of $2sN = 6$ connections per neurons. The optimal sparsity level does not correspond to a constant number of connections per neurons as it is not inversely proportional to the reservoir size, but follows a law in $N^{-0.38}$ (see Fig. \ref{fig: sparse}). At large reservoir sizes, more connections per neurons are required. We suspect that as large reservoirs are better approximations of the RK limit, it becomes more challenging to reduce $s$ while keeping the same approximation quality. 

\subsection{Optimal Deep Reservoir Computing sizes}

\begin{figure}
    \centering
    \includegraphics[width=\linewidth]{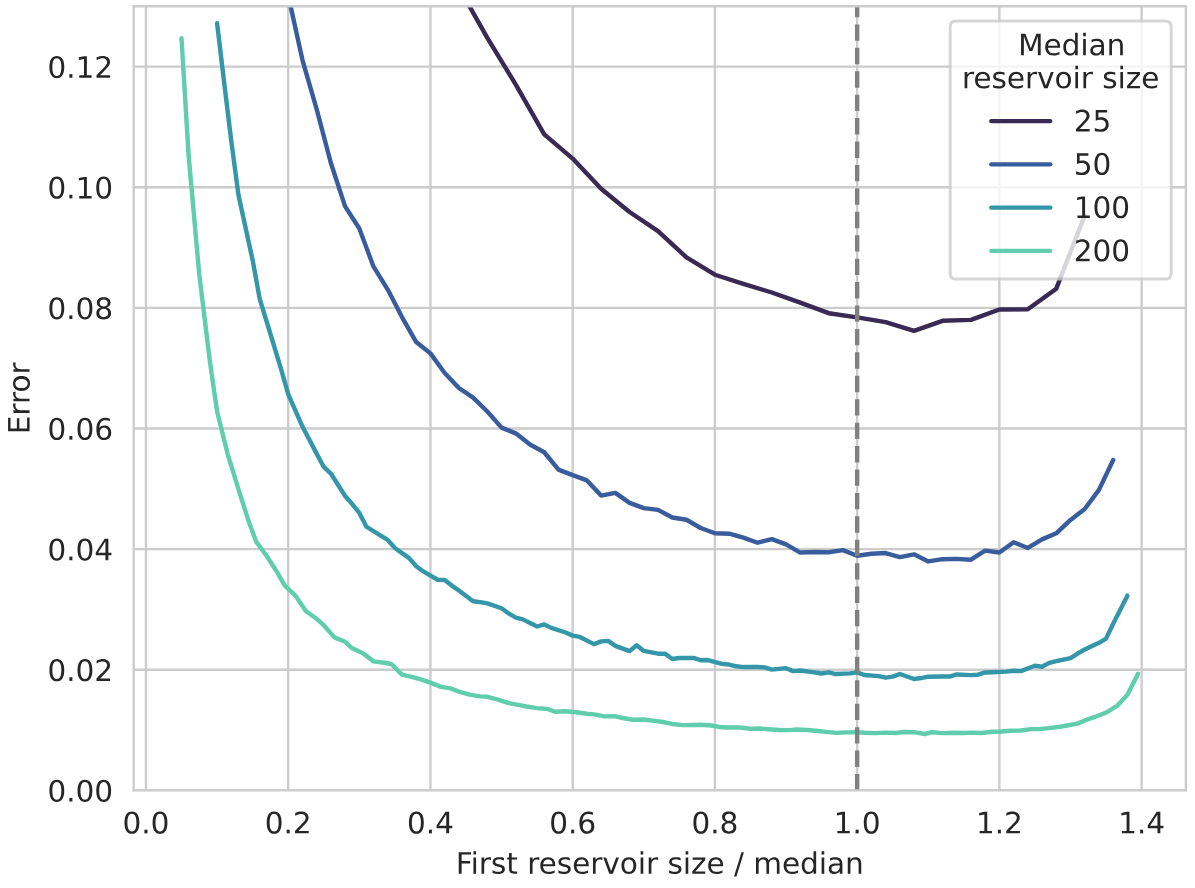}
    \caption{Error metric (Eq. \eqref{eq: metric convergence}) for Deep Reservoir Computing for varying reservoir sizes. We vary the first layer size $N_1$ for a fixed computational budget $N_1^2+N_2^2 = 2 \times N_\text{med}^2$ for different values of the median reservoir size $N_\text{med}$.}
    \label{fig: fig3 deep RC sizes}
\end{figure}

Our framework enables us to investigate how to optimally set the various reservoir weights in Deep Reservoir Computing, addressing the previously unresolved question of whether the first or second reservoir should be larger. To investigate this question, we compute the previous metric given in Eq. \eqref{eq: metric convergence} for a fixed computational budget $N_1^2 + N_2^2 = 2 \times N_\text{med}^2$ for different values of $N_\text{med}$. This quadratic scaling corresponds to the computational and memory complexity of the dense matrix multiplication, which is the limiting factor in Eq. \eqref{eq: RC update}. Each point is an average of 10'000 repetitions. 

The metric as a function of $N_1$ is depicted in Fig. \ref{fig: fig3 deep RC sizes}. We see that in general the extreme cases yield high error. When the first reservoir size is too small, the error of the first RK is detrimental even though the second reservoir is closer to its limit. Similarly, the second reservoir size cannot be too small or the second recurrent kernel is not well approximated. 

Between these extremes, there is a region for which the error is relatively small. For small computational budgets, this region is limited, both reservoirs need to be approximately the same size, while for large reservoirs, the actual reservoir sizes do not seem to matter as much as long as we avoid the extreme cases. We also observe that the minimal error is obtained for a first reservoir size slightly larger than the second. 

To obtain a quantitative answer, we performed a Nelder-Mead optimization to find the optimal reservoir sizes. For $L=2$, we obtain $n_1, n_2 = 209, 190$, for $L=3$, we obtain $n_1, n_2, n_3 = 207, 202, 190$. Thus, the optimal shapes have first reservoir sizes that are larger than subsequent ones. This decreases the noise that is transmitted to the next layers. 

In a nutshell, a good rule of thumb to choose the reservoir sizes in Deep Reservoir Computing is to choose them all equal. First reservoirs can be chosen slightly (around 5\%) larger than the last ones to decrease further the distance with the asymptotic limit performance. 

\subsection{Computational benchmark}

\begin{figure}
    \centering
    \includegraphics[width=\linewidth]{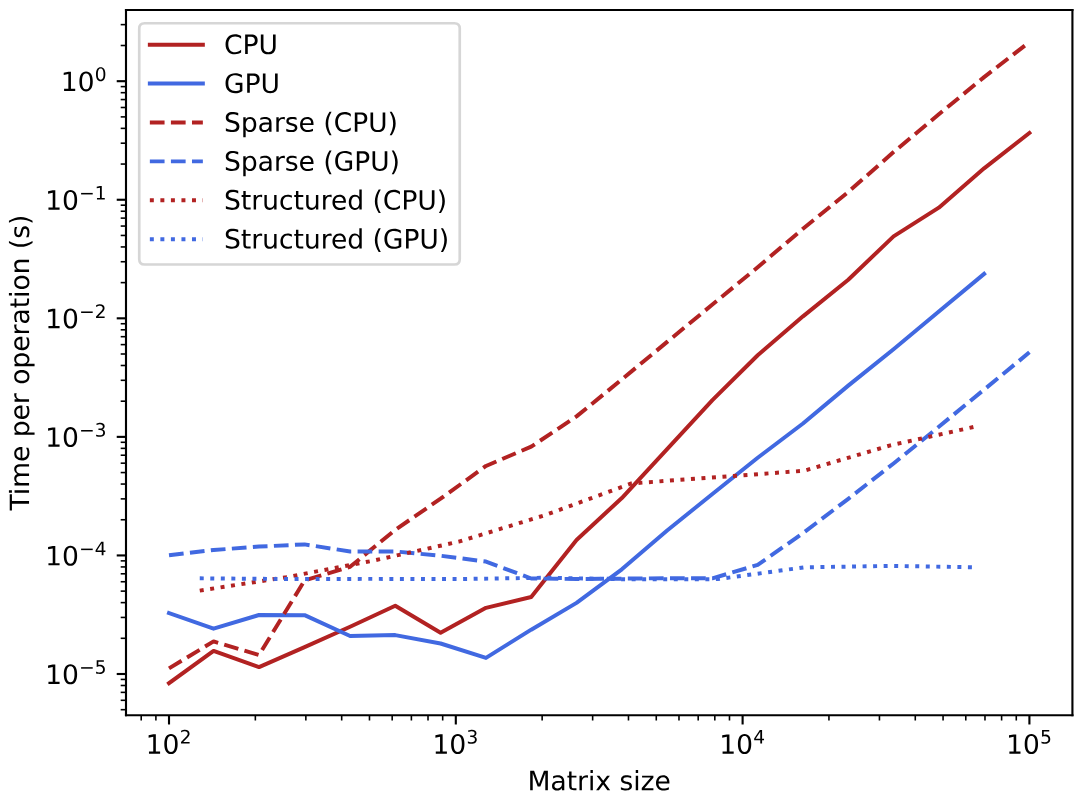}
    \caption{Time benchmark of the matrix-vector multiplication on CPU and GPU for Reservoir Computing, Sparse Reservoir Computing, and Dense Reservoir Computing. }
    \label{fig: fig5 benchmark}
\end{figure}

Since both RC, Sparse RC, and Structured RC converge to the same kernel limit, we perform a benchmark to provide guidelines on which topology to choose. We benchmark the matrix-vector multiplication between the square reservoir weight matrix $\Wrr$ and the reservoir state $\x$ since this is the bottleneck operation. This benchmark is performed on CPU (Intel Xeon 6240) and GPU (NVIDIA V100). The library used is crucial as low-level optimizations may impact performance. We use the Pytorch library for dense and sparse matrix multiplications and Fast Fourier Tranforms in the Structured RC case. Sparsity is fixed at $s=0.01$.

We first confirm that GPU acceleration is useful for large reservoir sizes, with the threshold being at a few hundred neurons for all techniques. CPUs, on the other hand, excel at small reservoir sizes.

For CPU-based computations, we find that sparse matrix operations in PyTorch are not particularly efficient. In contrast, vanilla RC is beneficial for small reservoir sizes while structured reservoir computing demonstrates superior performance as reservoir sizes increase. The results also demonstrate the quadratic computational complexity associated with both dense and sparse matrix multiplications, while the computational scaling of structured transforms is smaller.

In GPU-based computations, sparse matrix operations show improved efficiency compared to their CPU counterparts. However, Structured Reservoir Computing still outperforms both dense and sparse approaches. This indicates that for GPU implementations, structured transforms may offer the best performance, particularly for applications with reservoir sizes exceeding a few thousand nodes.

\section{Discussion}\label{sec:discussion}

In our study, we have derived the Recurrent Kernel limit of different reservoir topologies. We have shown that different topologies can lead to the same asymptotic limit. More specifically, the presence of sparsity does not affect convergence at all, which justifies the sparse initialization of reservoir weights to speed up computation. Convergence has been studied numerically and validated for a wide range of parameters, especially for bounded activation functions. Furthermore, we have derived how Recurrent Kernels extend to Deep Reservoir Computing, and how it sheds new insight on how to set the consecutive reservoir sizes. Finally, our timing benchmark has demonstrated the superior efficiency of Structured Reservoir Computing, particularly for implementations with large reservoir sizes.

The current study focused on theoretical convergence and speed comparison. It may be interesting to provide a benchmark on a particular Reservoir Computing task (to verify the equivalence of different topologies with the corresponding Recurrent Kernel limit) and optimize further for computational cost, e.g. through the use of a dedicated sparse linear algebra library. Other topologies based on random connections may also be explored theoretically such as: random features \cite{rahimi2007random} and extreme learning machines \cite{huang2006extreme}, 
which approximate a kernel in expectation with non-recursive random embeddings; gaussian processes \cite{mackay1998introduction}, stochastic processes defined by a mean and a covariance, which can give an accurate estimate of the uncertainty in regression tasks ; and random vector functional link networks \cite{malik2023random}, wherein the weights of the network are generated randomly and output weights are computed analytically.

\section*{Acknowledgements}

We would like to thank Rahul Parhi for insightful discussions and review of the paper. 
Giuseppe Alessio D'Inverno is partially funded by Indam GNCS group. Jonathan Dong is funded by the Swiss National Science Foundation (SNSF) under Grant PZ00P2\_216211.

\appendix

\bibliographystyle{elsarticle-harv} 
\bibliography{refs.bib}

\section{Iterable kernels for any rotationally-invariant distribution}

We prove here that the kernel limit defined in Eq. \eqref{eq: first kernel limit} is iterable as soon as the weight distribution $p(\w)$ is rotationally invariant. For any fixed timestep $t$, we can use this property to perform a change of basis:
\begin{equation}
    \begin{cases}
        \but = u_1 \be_1
        \\
        \bvt = v_1 \be_1 + v_2 \be_2
    \end{cases}
\end{equation}
with $\be_1$ and $\be_2$ the first two vectors of an orthonormal basis. Importantly, $u_1$, $v_1$, and $v_2$ only depend on scalar products $\|\but\|^2$, $\|\bvt\|^2$, and $\left(\but\right)^\top\bvt$:
\begin{equation}
    \begin{cases}
        u_1 = \|\but\|
        \\
        v_1 = \left(\bvt\right)^\top \be_1 = \frac{1}{\|\but\|} \left(\bvt\right)^\top \but
        \\
        v_2 = \sqrt{\|\bvt\|^2 - v_1^2} = \sqrt{\|\bvt\|^2 - \frac{1}{\|\but\|^2} \left(\left(\bvt\right)^\top \but\right)^2}
    \end{cases}
\end{equation}

The kernel limit in Eq. \eqref{eq: first kernel limit} can be rewritten as an integral over two gaussian random variables $w_1$ and $w_2$:
\begin{align}
    k_0\left(\but, \bvt\right) &= \int dw_1 dw_2 p(w_1) p(w_2) f(w_1 u_1) f(w_1 v_1 + w_2 v_2) \\
    &\equiv k\left(\|\but\|^2, \|\bvt\|^2, \left(\but\right)^\top \bvt\right),
\end{align}
since $u_1, v_1, v_2$ only depend on $\|\but\|^2, \|\bvt\|^2, \left(\but\right)^\top \bvt$.
Thus, the kernel limit for all RC algorithms with random gaussian weights is an iterable kernel, allowing us to define an associated Recurrent Kernel. As we see in this proof, it can be extended to any rotationally-invariant distribution of weights $p(\w)$. 

\section{Sparse Random Features}

We investigate the convergence of Eq. \eqref{eq: explicit scalar product} to its single-step limit defined in Eq. \eqref{eq: first kernel limit} when the weights $\w$ are sparse. This can be interpreted as a sparse Random Feature embedding. 

We initialize a random vector $\bu \in \mathbb{R}^d$ (i.i.d. uniform between 0 and 1) and generate Random Features embedding following:
\begin{equation}
    \psi(\bu) = f(\W \bu).
\end{equation}
$\W \in \mathbb{R}^{n\times d}$ is an i.i.d. random matrix and $f$ an element-wise non-linearity. In this context, we can also define the single-step kernel function $k_0(\bu, \bv)$ of Eq. \eqref{eq: first kernel limit}. The approximation error is given by:
\begin{equation}
    l = |\left(\psi(\bu)\right)^\top \psi(\bv) - k_0(\bu, \bv)|^2.
    \label{eq: loss sparse RF}
\end{equation}

\begin{figure}
    \centering
    \includegraphics[width=\linewidth]{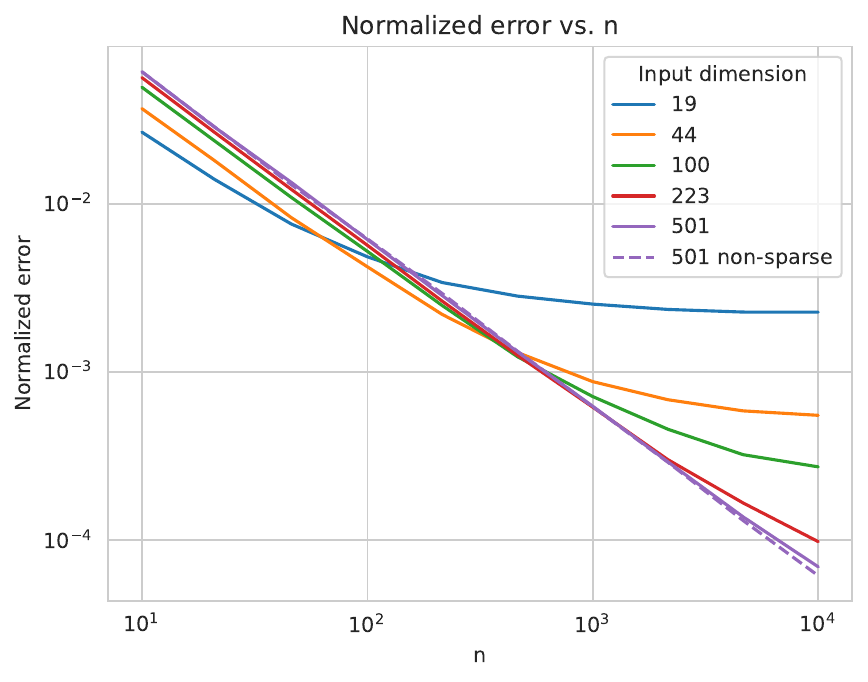}
    \caption{Approximation error Eq. \eqref{eq: loss sparse RF} of sparse Random Features as a function of Random Feature dimension $n$. An example of non-sparse convergence is given with dashed lines.}
    \label{fig:sparse app}
\end{figure}

This quantity is displayed in Fig. \ref{fig:sparse app} as a function of Random Feature dimension $n$, for different input dimension and for non-sparse (gaussian) and sparse ($s=0.1$) random vector $\w$. It is averaged over $10^4$ repetitions. 

First, we see that in the non-sparse case, convergence is achieved with a linear rate in $1/n$; only a single curve is displayed as the non-sparse curves only differ by a prefactor. In the sparse case, convergence is similar for small $n$, until a certain value after which the approximation error reaches a plateau. This shows that the sparsity level $s=0.1$ does not affect convergence of the Random Features to their kernel limit up that threshold on the output dimension $n$. This threshold varies greatly with the input dimension $d$. The greater the input dimension, the greater the number of random weights, and the less sparsity is affecting the convergence of sparse Random Features. 

In Reservoir Computing, input and output sizes $d$ and $n$ are similar, we see that this operating point is before this threshold for $s=0.1$. Instead, the sparsity level $s$ can be varied as displayed in Fig. \ref{fig: sparse}. 

\section{Derivation of the limit for leaky Reservoir Computing}

We motivate here the definition of the leaky Recurrent Kernel as defined in Eq. \eqref{eq: leaky RK definition}. 
Using the leaky RC update equation \eqref{eq: leaky RC definition}, we have:
\begin{align}
    \left(\xtt\right)^\top \ytt  &= \left(\frac{a}{\sqrt{N}} f\left(\W \but \right) + (1-a) \xt \right)^\top \\
    &\qquad\qquad \left(\frac{a}{\sqrt{N}} f\left(\W \bvt \right) + (1-a) \yt \right) \nonumber \\
    &= \frac{a^2}{N} f\left(\W \but \right)^\top f\left(\W \bvt \right) \\
    &\qquad\qquad \nonumber 
    + \frac{a(1-a)}{\sqrt{N}} f\left(\W \but \right)^\top \yt \\
    &\qquad\qquad \nonumber
    + \frac{a(1-a)}{\sqrt{N}} f\left(\W \bvt \right)^\top \xt \\
    &\qquad\qquad \nonumber
    + (1-a)^2 \left(\xt\right)^\top \yt
\end{align}

The first term converges to the non-sparse limit $k_0\left(\but, \bvt\right)$. The last term corresponds to the previous recurrent kernel limit $k^{(t)}(\bi^{(t-1)}, \bj^{(t-1)}, \ldots)$. Furthermore, we neglect the two cross-product terms, which results in Eq. \eqref{eq: leaky RK definition}. These cross-products are not straightforward to analyze as the previous reservoir state $\yt$ also depend on the random weights $\W$. 

\end{document}